\documentclass{article}

\usepackage[preprint]{aaj2026}
\usepackage[numbers]{natbib}

\usepackage[utf8]{inputenc} 
\usepackage[T1]{fontenc}    
\usepackage{hyperref}       
\usepackage{url}            
\usepackage{booktabs}       
\usepackage{newunicodechar}
\newunicodechar{└}{|}
\newunicodechar{─}{-}
\newunicodechar{├}{|}
\usepackage{fancyvrb}      
\usepackage{listings}       
\usepackage{listings}       
\usepackage{listings}       
\usepackage{listings}       
\lstdefinelanguage{yaml}{
  morekeywords={true,false,null,y,n},
  sensitive=false,
  morecomment=[l]{##},
  morestring=[b]"
}
\usepackage{amsmath}        
\usepackage{graphicx}       
\usepackage{amsfonts}       
\usepackage{nicefrac}       
\usepackage{microtype}      
\usepackage{xcolor}         

\title{EPOCH: An Agentic Protocol for Multi-Round System Optimization}

\author{%
  Zhanlin Liu \\
  ProRata.ai
\And
Yitao Li \\
ProRata.ai
}
\author{Zhanlin Liu \\ ProRata.ai \thanks{These authors contributed equally to this work} \and Yitao Li \\ ProRata.ai \footnotemark[1] \and Munirathnam Srikanth \\ ProRata.ai}

\begin{document}

\maketitle

\begin{abstract}
Autonomous agents are increasingly used to improve prompts, code, and machine learning systems through iterative execution and feedback. Yet existing approaches are usually designed as task-specific optimization loops rather than as a unified protocol for establishing baselines and managing tracked multi-round self-improvement. We introduce EPOCH, an engineering protocol for multi-round system optimization in heterogeneous environments. EPOCH organizes optimization into two phases: baseline construction and iterative self-improvement. It further structures each round through role-constrained stages that separate planning, implementation, and evaluation, and standardizes execution through canonical command interfaces and round-level tracking. This design enables coordinated optimization across prompts, model configurations, code, and rule-based components while preserving stability, reproducibility, traceability, and integrity of evaluation. Empirical studies in various tasks illustrate the practicality of EPOCH for production-oriented autonomous improvement workflows.
\end{abstract}

\section{Introduction}
Recent large language model systems can already decompose problems into executable plans and produce useful initial baselines for a wide range of reasoning, software engineering, and machine-learning tasks \cite{wang2023plan, yang2024swe}. In parallel, a growing body of work studies iterative self-improvement in task-specific settings, including self-improvement of language-model pipelines and prompt optimization in DSPy \cite{khattab2023dspy}, Promptbreeder \cite{fernando2309promptbreeder} and GEPA\cite{agrawal2025gepa}, hyperparameter search with AgentHPO \cite{liu2024large}, and autonomous software repair with RepairAgent \cite{bouzenia2024repairagent}. Together, these results suggest that large language model systems are increasingly capable not only of producing an initial solution but also of refining that solution through repeated interaction with tools, execution feedback, and evaluation signals.

However, these existing approaches are typically instantiated as task-specific optimization loops, domain-specific agents, or workflow frameworks rather than as a unified protocol to take a problem, establish an executable baseline, and then manage multi-round self-optimization under a consistent evaluation interface \cite{khattab2023dspy, agrawal2025gepa, liu2024large, bouzenia2024repairagent, yang2024swe}. As a result, they generally do not provide a shared protocol abstraction with explicit role separation, canonical execution interfaces, and round-level state tracking across heterogeneous optimization settings. At a high level, this missing layer can be understood as a protocolized decision loop, analogous to the observe-orient-decide-act (OODA) cycle proposed by Boyd for iterative decision-making in dynamic environments \cite{boyd2018discourse}. In such a loop, the system must observe the current state, reason about possible changes, act through constrained execution, and evaluate the result before proceeding. What is missing in existing agentic optimization systems is therefore not the ability to optimize individual artifacts, but a reproducible and deployment-oriented protocol for managing this loop across rounds and system components.

This gap matters in practical deployment environments, where performance rarely depends on a single artifact in isolation. Production machine-learning systems contain many interlocking components beyond model training, and system behavior is often shaped by interactions among code, configurations, data pipelines, and serving logic \cite{sculley2015hidden, xin2021production}. In such settings, improving a system may require coordinated changes between prompts, model parameters, rule systems, code logic, and data-processing stages, while maintaining reproducibility, evaluation integrity, and traceability for each optimization round \cite{sculley2015hidden, wolter2025more}.

In this work, we introduce EPOCH, an agentic protocol for multi-round system optimization in heterogeneous environments. EPOCH treats iterative improvement as a two-phase process: it first uses the model to analyze the problem, define an executable plan, and establish an initial baseline; it then performs structured multi-round self-optimization to improve the system over successive rounds. Rather than implementing a single optimizer for one artifact type, EPOCH provides a general orchestration layer that organizes optimization into role-constrained stages, separates hypothesis generation from evaluation, and standardizes execution through canonical command interfaces and round-level tracking.

EPOCH is designed for industry integration. The protocol supports configurable evaluation pipelines, heterogeneous optimization skills, and explicit tracking of system changes between rounds. We instantiate EPOCH across prompt tuning, hyperparameter tuning, rule-based optimization, and code improvement, and study how protocol-based orchestration supports stable, reproducible, and traceable optimization behavior across these task settings.

The main contributions of this work are:

\begin{itemize}
    \item EPOCH, an agentic protocol for transforming a problem specification into an executable baseline and then performing tracked multi-round self-optimization across heterogeneous improvement modalities.
    \item A two-phase orchestration framework that separates baseline establishment from iterative self-improvement and separates hypothesis generation, implementation, and evaluation within each optimization round.
    \item A unified protocol abstraction for industry-ready optimization workflows, with canonical execution interfaces, configurable evaluation pipelines, and round-level tracking of system changes.
    \item Task-specific instantiations and empirical studies that span prompt tuning, hyperparameter tuning, rule-based optimization, and code improvement show that protocol-based orchestration yields stable, reproducible, and traceable optimization trajectories in various task settings.
\end{itemize}

In addition, to support reproducibility, all the code, configurations, and instantiations of the task-specific protocol used in this work will be publicly released upon publication.

\section{Framework}

\begin{figure}
    \centering
    \includegraphics[width=0.95\linewidth]{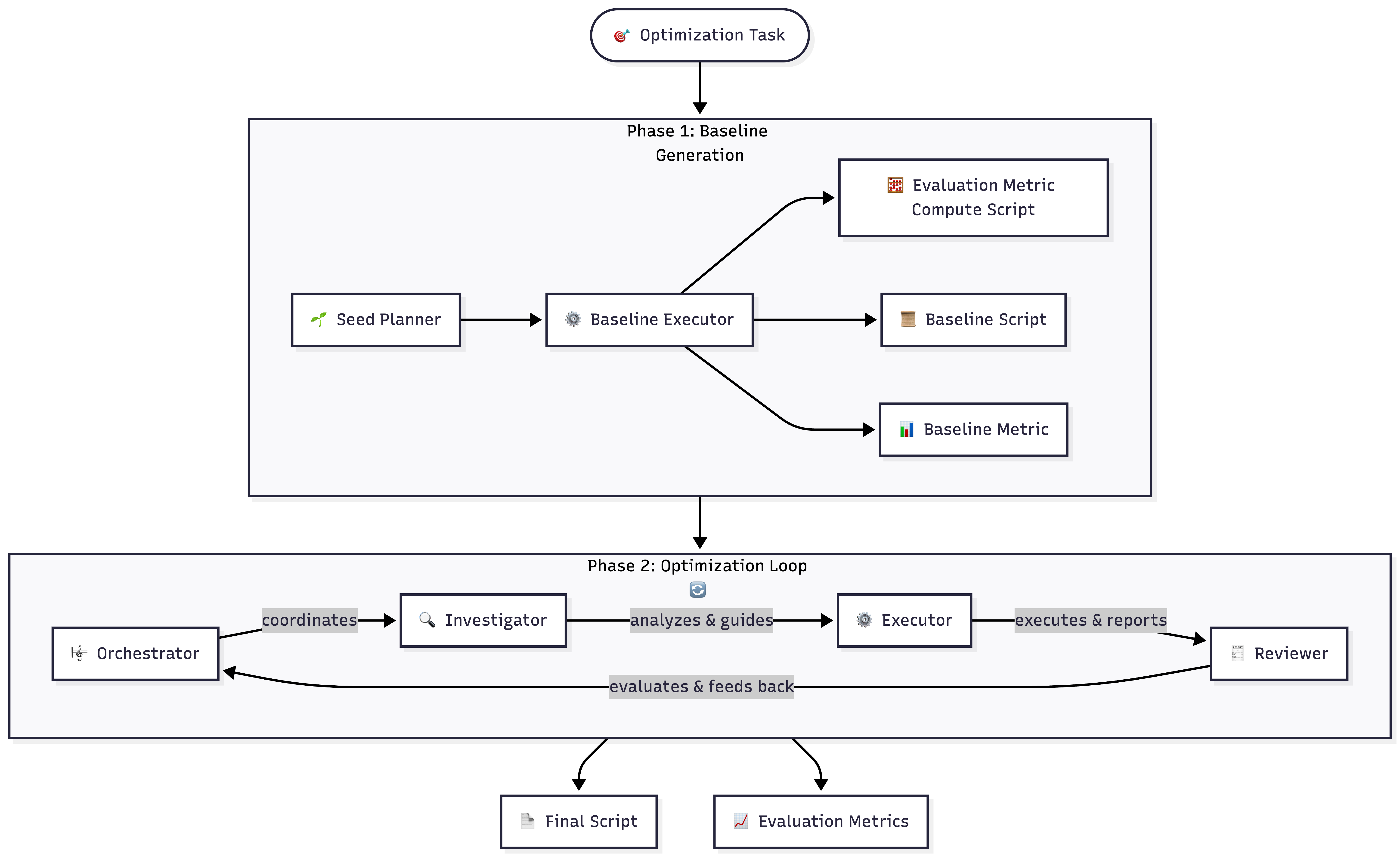}
    \caption{Architecture of EPOCH. EPOCH contains two phrases. Phrase 1 use seed planner and baseline executor to generate the evaluation metric compute script, baseline script, and baseline metric. Phrase 2 is a multi-round process, which improve the final optimization results through the orchestrator, investigator, executor, and reviewer. It generates the final script and evaluation metric at the end.}
    \label{fig:diagram}
\end{figure}

EPOCH treats iterative improvement as a \emph{protocol} rather than as a single optimizer. Given a problem specification, the protocol first establishes an executable baseline and then manages tracked multi-round self-improvement under a consistent evaluation interface. As shown in Figure ~\ref{fig:diagram}, its design is guided by three principles: (1) optimization is organized as a \textbf{two-phase process} consisting of baseline construction and iterative refinement; (2) each phase is decomposed into \textbf{role-constrained stages} that separate planning, implementation, and evaluation; and (3) all runs use \textbf{canonical execution interfaces and round-level tracking} so that changes can be reproduced, compared, and audited over time.

\subsection{Protocol Overview}

An EPOCH run begins from a \emph{problem specification} provided as either a natural-language task description or a structured task configuration. This specification defines the optimization target, evaluation interface, operational constraints, optimization budget, available data resources, and any task-specific runtime settings. From it, the EPOCH agent instantiates the appropriate protocol variant, determines the canonical execution and evaluation commands, and assigns role-specific permissions and data-access constraints.

At a high level, EPOCH organizes optimization as a sequence of explicit state transitions. The protocol first establishes an accepted baseline state and then coordinates successive rounds in which a candidate modification is proposed, implemented, evaluated under a standardized interface, and accepted or rejected. Each round produces structured artifacts, including metrics, round-to-round deltas, decisions, and supporting records of the actions taken. This turns optimization into a traceable history rather than an unstructured sequence of edits.

This control structure can also be understood through an OODA-style loop:  Observe, Orient, Decide, and Act, originally proposed as a model for iterative decision-making in dynamic environments \cite{boyd2018discourse}. EPOCH \emph{observes} the current accepted state and available evidence, \emph{orients} through task-specific analysis and protocol constraints, \emph{decides} on a candidate modification, and \emph{acts} by implementing and evaluating that modification. Unlike an informal decision loop, however, EPOCH makes each transition explicit through role separation, canonical interfaces, and round-level artifact tracking.

EPOCH is also customizable by design. Users may specify whether Phase~I (baseline construction) is required, whether Phase~II (multi-round self-improvement) is enabled, and which roles are instantiated in each phase. Roles may also be implemented differently across tasks; for example, the Reviewer may be an LLM-based critic or a purely metric-driven component. Similarly, round-level tracking may rely on local logs and artifacts or on native development records such as GitHub pull requests, review comments, and commit histories.

\subsection{Phase I: Baseline Construction}

Phase~I transforms a problem specification into an accepted executable baseline. In many practical settings, optimization does not begin from a validated system with a clean evaluation interface; instead, the initial state may consist only of a task description, partial code, or loosely defined requirements. EPOCH therefore treats baseline construction as part of the protocol rather than as an external prerequisite.

As shown in Figure~\ref{fig:diagram}, Phase~I typically contains two roles: \textbf{Seed Planner} and \textbf{Baseline Executor}. The Seed Planner analyzes the task specification and designs the initial system and evaluation interface, including execution entry points, expected inputs and outputs, and the minimal strategy needed for a valid baseline. The Baseline Executor converts this design into a runnable implementation, instantiates the required configuration and evaluation artifacts, and generates the first accepted metrics. If a validated baseline already exists, this phase can be reduced to a baseline-validation step rather than a full reconstruction.

\subsection{Phase II: Multi-Round Self-Improvement}

Once an accepted baseline has been established, EPOCH enters Phase~II, the multi-round self-improvement phase. Each round starts from the current accepted state and produces a candidate modification that is explicitly evaluated before being incorporated. The purpose of this phase is not merely to generate changes, but to ensure that changes are proposed and validated under a reproducible protocol.

As shown in Figure~\ref{fig:diagram}, Phase~II is organized around four logical roles: \textbf{Orchestrator}, \textbf{Investigator}, \textbf{Executor}, and \textbf{Reviewer}. The Orchestrator manages the round-level control flow, including budgets, tries, and transitions between accepted states. The Investigator analyzes the current system and available evidence to generate hypotheses for improvement under task-specific constraints, such as train-only access in machine-learning settings. The Executor implements the proposed changes within the admissible action space of the task, such as editing prompts, adjusting hyperparameters, modifying rules, or updating source code. The Reviewer then evaluates the candidate state under the canonical evaluation interface and determines whether it should be accepted.

A key property of this design is the separation between hypothesis generation, implementation, and evaluation. The component that decides whether a change is valid is not the same component that proposed or implemented it. This helps preserve the integrity of the evaluation, especially in settings where data leakage, overfitting, or inconsistent evaluation procedures would otherwise undermine the reliability of iterative optimization.

Although the logical structure of Phase~II remains fixed, its concrete realization is configurable. Some deployments may use an LLM-based reviewer, while others rely entirely on metric thresholds; some tasks may collapse or omit roles when the full structure is unnecessary. What remains constant is the protocol logic: each round begins from an accepted state, produces a constrained candidate modification, evaluates that candidate through a standardized interface, and records the decision together with its supporting artifacts. In this way, EPOCH turns iterative improvement into a sequence of explicit and auditable transitions that support stability, traceability, and reproducibility throughout the optimization trajectory.

\subsection{Round-Level Tracking and Evaluation Integrity}

EPOCH records each optimization round as a structured unit of experimentation. Each round is associated with a candidate change, the evidence motivating that change, the resulting evaluation metrics, and the final accept/reject decision. This round-level structure allows us to reconstruct how and why the system evolved over time.

To preserve comparability across rounds, EPOCH standardizes execution through canonical command interfaces derived from the task specification. This reduces evaluation drift by ensuring that improvements are measured under a shared interface rather than under ad hoc scripts or changing procedures. For deterministic tasks, evaluation may rely on committed result artifacts; for non-deterministic tasks, evaluation may be re-run each round while still recording all inputs, outputs, and decisions.

More broadly, EPOCH is designed to preserve \textbf{the integrity} of the evaluation. The roles that generate or implement hypotheses are distinct from the role that evaluates them. Where appropriate, the protocol also enforces train/evaluation separation, leakage checks, and task-specific acceptance constraints. These mechanisms are especially important in deployment-oriented settings, where iterative optimization must remain reproducible, auditable, and safe to integrate into existing workflows \cite{sculley2015hidden, wolter2025more}.

\subsection{Task-Specific Protocol Instantiations}

EPOCH is a general orchestration layer whose behavior is specialized through task-specific protocol instantiations. In our current implementation, we study four representative settings: \textbf{prompt tuning}, \textbf{hyperparameter fine-tuning}, \textbf{rule-based optimization}, and \textbf{code improvement}. These correspond to classes of iterative optimization that have often been studied through separate task-specific systems \cite{khattab2023dspy, fernando2309promptbreeder, agrawal2025gepa, liu2024large, bouzenia2024repairagent}. EPOCH does not replace these methods with a single optimizer; instead, it provides a shared protocol in which heterogeneous optimization tasks can be carried out and tracked consistently.

In \textbf{code improvement}, the optimized artifact is the program logic. Unlike the other settings, this task often uses a fully visible test suite rather than a train/evaluation split. EPOCH therefore treats it as a staged optimization problem: first achieving correctness, then improving execution time or memory while preserving correctness. Even in this setting, the same protocol principles remain in place: analysis, implementation, evaluation, and tracking are kept distinct.

In \textbf{hyperparameter fine-tuning}, the optimized artifacts are tunable training hyperparameters, while the base model and architecture remain fixed. The Investigator analyzes training dynamics and failure patterns, the Executor updates the hyperparameter configuration, and the Reviewer evaluates whether held-out performance improves without unacceptable overfitting.

In \textbf{prompt tuning}, the optimized artifact is the prompt, while the underlying language model remains fixed. The Investigator analyzes failures in the training split, the Executor updates instruction or example files, and the Reviewer evaluates stalled performance while checking for leakage of the evaluation.

In \textbf{rule-based optimization}, the optimized artifact is a set of rules, such as thresholds, precedence relations, or symbolic conditions. The Investigator identifies coverage gaps, false positives, false negatives, and precedence conflicts, the Executor modifies the rules, and the Reviewer evaluates whether the revised rules improve the target metric without becoming overly specific or brittle.

Together, these instantiations show that EPOCH is not tied to a single modality. The same protocol can support a range of optimization tasks by varying the admissible action space, evidence constraints, and task-specific decision logic while preserving a common orchestration structure.

\subsection{Protocol Variants and Extensibility}

Although EPOCH defines a common protocol, it is not intended to be rigid. Different deployment environments may require different variants of the protocol. For example, Phase~I may be unnecessary when a validated baseline and evaluation interface already exist; the Reviewer may be implemented as a lightweight metric-monitoring component rather than a separate review stage; and version-control actions may be disabled in local-only workflows. Similarly, some tasks rely on deterministic committed-result comparison, whereas others require repeated evaluation because execution is stochastic.

To support such variation, EPOCH includes a meta-level skill construction mechanism that generates task-specific protocol instantiations from a free-form optimization goal. This allows the core protocol structure to remain stable while adapting role composition, constraints, and artifacts to the needs of a given task. EPOCH should therefore be understood as a configurable \emph{protocol family} for industry-oriented iterative improvement rather than as a single fixed pipeline.

\section{Empirical Evaluation of EPOCH Across Skills}

We evaluate EPOCH across four task-specific skill instantiations: \textbf{code improvement}, \textbf{hyperparameter fine-tuning}, \textbf{prompt tuning}, and \textbf{rule-based optimization}. These settings span heterogeneous optimization targets, including source code, prompts, model configurations, and symbolic rules. Across all settings, the underlying protocol remains the same: EPOCH establishes or validates a baseline, performs structured investigation, applies constrained modifications, and accepts or rejects each round under a standardized evaluation interface.

Our goal is not to present EPOCH as a state-of-the-art task-specific optimizer. Instead, these experiments evaluate EPOCH as a \emph{protocol layer} for organizing iterative improvement. We therefore use controlled task settings that make round-level behavior observable and allow us to study baseline construction, iterative refinement, rejection and retry behavior, evaluation integrity, and self-termination under different operational constraints.

\subsection{Experimental Setup}

All experiments follow the same high-level protocol. Each run begins from a task specification that defines the optimization target, the maximum number of rounds, any retry limits, the evaluation metric, and task specific constraints such as train/evaluation separation, visible tests, or overfitting thresholds. For tasks with held-out evaluation data, the Investigator is restricted to the training split and the Reviewer makes accept/reject decisions on the evaluation split. For deterministic tasks such as code optimization and rule-based classification, EPOCH can compare committed result artifacts directly; for non-deterministic tasks such as prompt optimization, evaluation may be re-run each round while retaining the same protocol structure.

The four experiments differ in their target optimization and admissible actions. In code improvement, the system optimizes the program logic under a visible test suite and transitions from correctness to performance once all tests pass. In hyperparameter fine-tuning, the system adjusts only declared hyperparameters while keeping the model architecture fixed. In prompt tuning, the system modifies prompt artifacts while keeping the underlying language model fixed and enforcing leakage checks. In rule-based optimization, the system refines symbolic thresholds and precedence relations under train/evaluation separation. Together, these experiments test whether a shared protocol can support heterogeneous optimization workflows without changing its core control structure.

\section{Empirical Evaluation of EPOCH Across Skills}

We evaluate EPOCH across four task-specific skill instantiations: \textbf{code improvement}, \textbf{hyperparameter fine-tuning}, \textbf{prompt tuning}, and \textbf{rule-based optimization}. These settings span heterogeneous optimization targets, including source code, prompts, model configurations, and symbolic rules. Across all settings, the underlying protocol remains the same: EPOCH establishes or validates a baseline, performs structured investigation, applies constrained modifications, and accepts or rejects each round under a standardized evaluation interface.

Our goal is not to present EPOCH as a state-of-the-art task-specific optimizer. Instead, these experiments evaluate EPOCH as a \emph{protocol layer} for organizing iterative improvement. We therefore use controlled task settings that make round-level behavior observable and allow us to study baseline construction, iterative refinement, rejection and retry behavior, evaluation integrity, and self-termination under different operational constraints.

\subsection{Experimental Setup}

All experiments follow the same high-level protocol. Each run begins from a task specification that defines the optimization target, the maximum number of rounds, any retry limits, the evaluation metric, and task specific constraints such as train/evaluation separation, visible tests, or overfitting thresholds. For tasks with held-out evaluation data, the Investigator is restricted to the training split and the Reviewer makes accept/reject decisions on the evaluation split. For deterministic tasks such as code optimization and rule-based classification, EPOCH can compare committed result artifacts directly; for non-deterministic tasks such as prompt optimization, evaluation may be re-run each round while retaining the same protocol structure.

The four experiments differ in their target optimization and admissible actions. In code improvement, the system optimizes the program logic under a visible test suite and transitions from correctness to performance once all tests pass. In hyperparameter fine-tuning, the system adjusts only declared hyperparameters while keeping the model architecture fixed. In prompt tuning, the system modifies prompt artifacts while keeping the underlying language model fixed and enforcing leakage checks. In rule-based optimization, the system refines symbolic thresholds and precedence relations under train/eval separation. Together, these experiments test whether a shared protocol can support heterogeneous optimization workflows without changing its core control structure.

\subsection{Code Improvement: Fibonacci Calculator}

We evaluate EPOCH in a deterministic code-optimization setting using a Fibonacci CLI calculator that must compute $\mathrm{fib}(n)$ for large $n$ (up to $10^6$) within a 2-second wall-clock budget. The task was configured as \texttt{code\_improvement}, with a minimum improvement threshold of 5\% per performance round and a maximum of 5 optimization rounds. Unlike the training / evaluation settings used in the other skills, all 19 tests were visible throughout the optimization. These tests covered correctness, invalid-input handling, and performance constraints. EPOCH therefore treated the task as a two-stage optimization problem: first achieving full correctness, then improving runtime while preserving a 100\% \ test pass rate.

\begin{table}[t]
  \centering
  \caption{Round-level progression for code improvement on the Fibonacci calculator. Execution times are median values over 10 runs.}
  \label{tab:fibonacci-progression}
  \begin{tabular}{clcccc}
    \toprule
    Round & Key change & Tests Passed & $\mathrm{fib}(10^5)$ & $\mathrm{fib}(10^6)$ & Verdict \\
    \midrule
    1 & Iterative $O(n)$ baseline & 17/19 & 94\,ms & 8{,}420\,ms & Baseline \\
    2 & Replace loop with fast doubling & 19/19 & 1.00\,ms & 34.3\,ms & Accept \\
    3 & Use \texttt{gmpy2} \texttt{mpz} arithmetic & 19/19 & 0.16\,ms & 2.39\,ms & Accept \\
    4 & Replace Python logic with \texttt{gmpy2.fib} & 19/19 & \textbf{0.07\,ms} & \textbf{1.33\,ms} & Accept \\
    5 & No meaningful further gain detected & 19/19 & --- & --- & Terminate \\
    \bottomrule
  \end{tabular}
\end{table}

As shown in Table~\ref{tab:fibonacci-progression}, EPOCH first resolved the failing performance tests by replacing the naive iterative implementation with a fast-doubling algorithm, which reduced the computation from linear iteration to logarithmic recursion and immediately brought the system to full correctness under the visible test suite. Once all tests passed, the protocol automatically transitioned to performance optimization. The following accepted rounds moved from algorithmic improvement to optimized big-integer arithmetic and finally to direct use of GMP's native Fibonacci routine. The run terminated when profiling showed that nearly all remaining execution time was already inside the native implementation, and no further Python-level change could plausibly satisfy the configured improvement threshold. This experiment shows that EPOCH can manage staged optimization, automatically switch objectives once correctness is saturated, and self-terminate when further improvement is no longer meaningful.

\subsection{Hyperparameter Fine-Tuning: MNIST Classification}

We evaluated EPOCH on a constrained hyperparameter tuning task using a pretrained MobileNetV2~\cite{sandler2018mobilenetv2} classifier on MNIST~\cite{lecun1998gradient}. The task was configured as \texttt{fine-tuning}, with a minimum improvement threshold of 2\% (\texttt{min\_delta}${}=0.02$), a maximum of 3 rounds and 1 retry per round. Only the classifier head was trained; all convolutional layers remained frozen. To make hyperparameter choice consequential, we used a small deterministic subset consisting of 200 training images and 60 evaluation images, and limited training to 3 epochs. The optimizer could modify only two declared hyperparameters: optimizer choice (\texttt{adam}, \texttt{adamw}, \texttt{sgd}) and learning rate in the range \texttt{[0.0001, 0.01]}. Train/ test separation was enforced throughout, together with an overfit constraint requiring a train-test gap $< 0.15$.

\begin{table}[t]
  \centering
  \caption{Round-level progression for hyperparameter fine-tuning on MNIST. All runs are deterministic (seed=42).}
  \label{tab:mnist-progression}
  \begin{tabular}{clcccc}
    \toprule
    Round & Key change & Train Acc. & Eval Acc. & Gap & Verdict \\
    \midrule
    1 & Adam, $\text{lr}=0.001$ & 0.6050 & 0.5333 & 0.072 & Baseline \\
    2 & AdamW, $\text{lr}=0.005$ & 0.7100 & 0.6167 & 0.093 & Accept \\
    3 & AdamW, $\text{lr}=0.010$ & 0.6700 & 0.5500 & 0.120 & Reject \\
    3R & SGD, $\text{lr}=0.008$, momentum $0.9$ & 0.7050 & \textbf{0.6667} & 0.038 & Accept \\
    \bottomrule
  \end{tabular}
\end{table}

As shown in Table~\ref{tab:mnist-progression}, EPOCH first diagnosed an underfitting of the baseline loss trajectory and improved hold-out precision by switching from Adam to AdamW while increasing the learning rate. A more aggressive learning-rate increase then caused instability and was rejected, showing that the protocol does not simply accept monotonic parameter escalation. In retry, EPOCH changed strategy rather than magnitude, switching optimizer family to SGD with momentum, and recovering the best final evaluation accuracy while also reducing the train-eval gap. This experiment illustrates three protocols behaviors: strict train/eval separation during investigation and review, explicit rejection, and retry handling when a candidate change degrades performance, and traceable round-level optimization under deterministic execution.

\subsection{Prompt Tuning: SST-2 Sentiment Classification}

We evaluate EPOCH in a prompt-optimization setting where the underlying model is fixed and only prompt artifacts may be modified. We apply the protocol to binary sentiment classification on SST-2~\cite{socher2013recursive} using \texttt{gpt-4.1-nano}. The task was configured as \texttt{prompt\_tune}, with a minimum improvement threshold of 2\%, a maximum of 5 rounds, and strict separation between training and evaluation. To keep the round-level behavior interpretable, we used a small balanced subset of 20 training examples and 12 holdout evaluation examples, sampled deterministically from SST-2. All prompt modifications were derived exclusively from training failures, while evaluation examples were reserved for accept/reject decisions. This experiment is intended as a protocol-level demonstration of iterative prompt refinement under leakage constraints rather than as a competitive benchmark on SST-2.

\begin{table}[t]
  \centering
  \caption{Round-level progression for prompt tuning on SST-2 using \texttt{gpt-4.1-nano}. Train ($n{=}20$) and eval ($n{=}12$) subsets are disjoint.}
  \label{tab:sentiment-progression}
  \begin{tabular}{clcccc}
    \toprule
    Round & Key change & Train Acc. & Eval Acc. & Verdict \\
    \midrule
        1 & Generic zero-shot system prompt & 0.8000 & 0.8333 & Baseline \\
    2 & Add domain framing and negation guidance & 0.8500 & 0.9167 & Accept \\
    3 & Add 6 training-derived few-shot examples & \textbf{0.9000} & \textbf{1.0000} & Accept \\
    \bottomrule
  \end{tabular}
\end{table}

As shown in Table~\ref{tab:sentiment-progression}, EPOCH first improved performance by refining the system prompt to better reflect SST-2's movie-review domain and common failure modes such as fragment ambiguity and negation. Then a small number of demonstrations were introduced from a few-shots drawn exclusively from the training set to resolve the remaining errors. In all rounds, the leakage check passed, confirming that no evaluation examples were incorporated into the prompt. The protocol reached a perfect held-out accuracy in three rounds and was terminated early rather than exhaust the full optimization budget. This experiment shows that EPOCH can support autonomous prompt refinement while preserving evaluation integrity through train/evaluation separation, leakage checks, and controlled stopping behavior.

\subsection{Rule-Based Optimization: Iris Classification}

We evaluate EPOCH on a symbolic optimization task in which the optimized artifact is an interpretable rule set. We apply the protocol to classifying Iris species~\cite{fisher1936use} using hand-crafted decision rules on sepal length, sepal width, petal length, and petal width. The task was configured as \texttt{rule\_based}, with a minimum improvement threshold of 1\%, a maximum of 4 optimization rounds and 2 retries per round. The data set was deterministically split into 105 training samples and 45 evaluation samples. As in the other non-code tasks, failure analysis was restricted to the training split, and acceptance decisions were made on the held-out evaluation split.

\begin{table}[t]
  \centering
  \caption{Round-level progression for rule-based optimization on Iris. Train ($n{=}105$) and eval ($n{=}45$) subsets are disjoint.}
  \label{tab:iris-progression}
  \begin{tabular}{clcccc}
    \toprule
    Round & Key change & Train Acc. & Eval Acc. & Verdict \\
    \midrule
    1 & Petal-based baseline rules & 0.9524 & 0.9778 &  Baseline \\
    2 & Add sepal-width boundary condition & 0.9619 & 1.0000 &  Accept \\
    3 & Threshold refinement & 0.9810 & 1.0000 & No eval gain \\
    4 & Ratio-based exception rule & \textbf{0.9905} & \textbf{1.0000}  & Reject \\
    \bottomrule
  \end{tabular}
\end{table}

As shown in Table~\ref{tab:iris-progression}, EPOCH first improved the ruleset through boundary refinement at the versicolor--virginica decision boundary, reaching perfect held-out accuracy in the second round. It then explored additional train-side refinements, but because the evaluation metric was already saturated, those increasingly specific modifications did not produce measurable hold-out gain. The protocol therefore rejected the final ratio-based exception and terminated the run rather than continuing to refine for training-only benefit. This experiment shows that EPOCH can structure optimization for symbolic systems and enforce evaluation discipline even when the target is not a learned model.

\subsection{Cross-Skill Findings}

Across all four tasks, EPOCH preserves the same protocol logic while adapting to very different optimization settings. Code improvement operates over a fully visible test suite and uses staged objectives; hyperparameter tuning relies on train/evaluation separation and overfitting guards; prompt tuning emphasizes leakage prevention and non-deterministic evaluation; and rule-based optimization focuses on interpretable symbolic logic. Despite these differences, each run follows the same structure: baseline establishment or validation, structured investigation, constrained implementation, evidence-based review, and round-level tracking.

Several common behaviors emerge across the experiments. First, EPOCH consistently turns optimization into a sequence of explicit state transitions, making it possible to reconstruct why each accepted or rejected change occurred. Second, the protocol supports rejection and retry behavior without breaking the overall control structure, as seen most clearly in hyperparameter tuning. Third, EPOCH can terminate optimization when the configured criteria indicate diminishing returns, whether because the system has reached a practical performance ceiling, as in Fibonacci, or because the held-out metric is already saturated, as in SST-2 and Iris. Taken together, these experiments support the central claim of the paper: a shared protocol layer can coordinate heterogeneous improvement workflows while preserving stability, reproducibility, traceability, and evaluation integrity.

\section{Conclusion and Future Work}

We introduced EPOCH, an agentic protocol for transforming a problem specification into an executable baseline and then managing tracked multi-round self-improvement. Rather than proposing a new task-specific optimizer, EPOCH contributes a protocol layer for organizing iterative improvement as a reproducible engineering process. By separating baseline construction from iterative refinement, decomposing each round into roles-constrained stages, and standardizing execution through canonical interfaces and round-level tracking, EPOCH provides a unified framework for stable, traceable, and evaluation-consistent optimization across heterogeneous tasks.

This protocol-oriented perspective is particularly relevant in deployment-facing environments, where iterative improvement must be reproducible, auditable, and safe to integrate into existing workflows. Across prompt tuning, hyperparameter fine-tuning, rule-based optimization, and code improvement, our results show that a shared protocol can support diverse optimization modalities while preserving a common structure for experimentation, review, and decision-making.

An important direction for future work is extending EPOCH from single-target iterative optimization to multi-agent coordination over complex systems. Many practical optimization problems involve multiple interacting subsystems, including training pipelines, inference-time prompting strategies, model-serving deployments, monitoring layers, and downstream application logic. In such settings, improvement may require several specialized agents operating in parallel or hierarchically across different parts of the system, with the protocol coordinating their actions through shared evaluation checkpoints and system-level metrics. Extending EPOCH in this direction would move agentic optimization from artifact-level refinement toward protocolized coordination of full production systems.

\bibliography{biblio.bib}

@article{fisher1936use,
    title   = {The use of multiple measurements in taxonomic problems},
    author  = {Fisher, Ronald A},
    journal = {Annals of Eugenics},
    volume  = {7},
    number  = {2},
    pages   = {179--188},
    year    = {1936}
  }

@inproceedings{sandler2018mobilenetv2,
    title     = {MobileNetV2: Inverted Residuals and Linear Bottlenecks},
    author    = {Sandler, Mark and Howard, Andrew and Zhu, Menglong and Zhmoginov, Andrey and Chen, Liang-Chieh},
    booktitle = {Proceedings of the IEEE Conference on Computer Vision and Pattern Recognition},
    pages     = {4510--4520},
    year      = {2018}
  }

@article{lecun1998gradient,
    title   = {Gradient-based learning applied to document recognition},
    author  = {LeCun, Yann and Bottou, L{\'e}on and Bengio, Yoshua and Haffner, Patrick},
    journal = {Proceedings of the IEEE},
    volume  = {86},
    number  = {11},
    pages   = {2278--2324},
    year    = {1998}
  }

@inproceedings{socher2013recursive,
    title     = {Recursive Deep Models for Semantic Compositionality Over a Sentiment Treebank},
    author    = {Socher, Richard and Perelygin, Alex and Wu, Jean and Chuang, Jason and Manning, Christopher D. and Ng, Andrew and Potts, Christopher},
    booktitle = {Proceedings of the 2013 Conference on Empirical Methods in Natural Language Processing},
    pages     = {1631--1642},
    year      = {2013}
  }

@article{agrawal2025gepa,
  title={Gepa: Reflective prompt evolution can outperform reinforcement learning},
  author={Agrawal, Lakshya A and Tan, Shangyin and Soylu, Dilara and Ziems, Noah and Khare, Rishi and Opsahl-Ong, Krista and Singhvi, Arnav and Shandilya, Herumb and Ryan, Michael J and Jiang, Meng and others},
  journal={arXiv preprint arXiv:2507.19457},
  year={2025}
}

@article{khattab2023dspy,
  title={Dspy: Compiling declarative language model calls into self-improving pipelines},
  author={Khattab, Omar and Singhvi, Arnav and Maheshwari, Paridhi and Zhang, Zhiyuan and Santhanam, Keshav and Vardhamanan, Sri and Haq, Saiful and Sharma, Ashutosh and Joshi, Thomas T and Moazam, Hanna and others},
  journal={arXiv preprint arXiv:2310.03714},
  year={2023}
}

@inproceedings{wang2023plan,
  title={Plan-and-solve prompting: Improving zero-shot chain-of-thought reasoning by large language models},
  author={Wang, Lei and Xu, Wanyu and Lan, Yihuai and Hu, Zhiqiang and Lan, Yunshi and Lee, Roy Ka-Wei and Lim, Ee-Peng},
  booktitle={Proceedings of the 61st annual meeting of the association for computational linguistics (volume 1: long papers)},
  pages={2609--2634},
  year={2023}
}

@article{yang2024swe,
  title={Swe-agent: Agent-computer interfaces enable automated software engineering},
  author={Yang, John and Jimenez, Carlos E and Wettig, Alexander and Lieret, Kilian and Yao, Shunyu and Narasimhan, Karthik and Press, Ofir},
  journal={Advances in Neural Information Processing Systems},
  volume={37},
  pages={50528--50652},
  year={2024}
}

@article{liu2024large,
  title={Large language model agent for hyper-parameter optimization},
  author={Liu, Siyi and Gao, Chen and Li, Yong},
  journal={arXiv preprint arXiv:2402.01881},
  year={2024}
}

@article{fernando2309promptbreeder,
  title={Promptbreeder: Self-referential self-improvement via prompt evolution (2023)},
  author={Fernando, Chrisantha and Banarse, Dylan and Michalewski, Henryk and Osindero, Simon and Rockt{\"a}schel, Tim},
  journal={arXiv preprint arXiv:2309.16797},
  year={2023}
}

@article{bouzenia2024repairagent,
  title={Repairagent: an autonomous, llm-based agent for program repair.(2024)},
  author={Bouzenia, Islem and Devanbu, Premkumar and Pradel, Michael},
  journal={arXiv preprint arXiv:2403.17134},
  year={2024}
}

@article{sculley2015hidden,
  title={Hidden technical debt in machine learning systems},
  author={Sculley, David and Holt, Gary and Golovin, Daniel and Davydov, Eugene and Phillips, Todd and Ebner, Dietmar and Chaudhary, Vinay and Young, Michael and Crespo, Jean-Francois and Dennison, Dan},
  journal={Advances in neural information processing systems},
  volume={28},
  year={2015}
}

@inproceedings{xin2021production,
  title={Production machine learning pipelines: Empirical analysis and optimization opportunities},
  author={Xin, Doris and Miao, Hui and Parameswaran, Aditya and Polyzotis, Neoklis},
  booktitle={Proceedings of the 2021 international conference on management of data},
  pages={2639--2652},
  year={2021}
}

@article{wolter2025more,
  title={More Rigorous Software Engineering Would Improve Reproducibility in Machine Learning Research},
  author={Wolter, Moritz and Veeramacheneni, Lokesh and Hoyt, Charles Tapley},
  journal={arXiv preprint arXiv:2502.00902},
  year={2025}
}

@book{boyd2018discourse,
  title={A discourse on winning and losing},
  author={Boyd, John R and others},
  volume={400},
  year={2018},
  publisher={Air University Press Maxwell Air Force Base, AL}
}


\appendix

\section{Appendix}

\subsection{Role Definitions}
\label{app:roles}

EPOCH decomposes optimization into a small set of logical roles. These roles define \emph{protocol functions} rather than fixed agent implementations, and may be realized by an LLM, a scripted component, or a hybrid mechanism depending on the task. Not every task requires every role in full form: Phase~I may be reduced to baseline validation when a runnable system already exists, and the Reviewer may be implemented either as an LLM critic or as a lightweight metric-driven component.

What remains invariant is the protocol logic: initialization establishes an accepted baseline, investigation proposes a candidate improvement, execution realizes that candidate within the admissible action space, and review determines whether it becomes the next accepted state. This separation of functions is the main mechanism by which EPOCH preserves reproducibility, traceability, and evaluation integrity across rounds.

\subsubsection{Seed Planner}

\textbf{Purpose.} The Seed Planner transforms a problem specification into an initial design for the baseline system and its evaluation interface.

\textbf{Inputs.} Problem specification, task type, available data description, optimization goal, and any runtime constraints provided at initialization.

\textbf{Outputs.} A baseline design describing the expected file structure, execution entry points, required inputs and outputs, evaluation interface, and the minimal strategy needed to produce a valid initial system.

\textbf{Constraints.} The Seed Planner is restricted to planning and design. It does not directly implement artifacts, execute commands, or make accept/reject decisions. In tasks with protected evaluation data, it does not inspect held-out evaluation examples.

\subsubsection{Baseline Executor}

\textbf{Purpose.} The Baseline Executor converts the Seed Planner's design into an executable baseline and generates the initial metrics required for subsequent optimization rounds.

\textbf{Inputs.} Baseline design, task configuration, execution environment, and any required resources such as code scaffolds, datasets, or model configuration.

\textbf{Outputs.} A runnable baseline implementation, the evaluation entry point, any required configuration artifacts, and the first accepted metrics artifact.

\textbf{Constraints.} The Baseline Executor is responsible for initialization rather than iterative improvement. Its goal is to establish a valid starting state, not to optimize beyond the baseline requirements.

\subsubsection{Orchestrator}

\textbf{Purpose.} The Orchestrator manages round-level control flow during multi-round optimization.

\textbf{Inputs.} Accepted system state, task specification, optimization budget, retry limits, and the artifacts generated in prior rounds.

\textbf{Outputs.} Round initialization, delegation to downstream roles, round-level state transitions, and the final optimization trajectory linking accepted and rejected candidates over time.

\textbf{Constraints.} The Orchestrator does not directly implement changes or decide round outcomes based on its own preferences. Its function is to govern sequencing, enforce protocol budgets and role constraints, and record transitions between accepted states.

\subsubsection{Investigator}

\textbf{Purpose.} The Investigator analyzes the current accepted state and available evidence to generate hypotheses for improvement.

\textbf{Inputs.} Accepted system state, training-side signals or visible test results, previous round artifacts, and any task-specific investigation budget such as sample limits.

\textbf{Outputs.} An investigation artifact describing observed failure patterns, bottlenecks, rationale, and a proposed modification strategy for the current round.

\textbf{Constraints.} The Investigator does not directly implement changes or make final accept/reject decisions. Its access is limited by the task specification. For example, it may be restricted to the training split in machine-learning tasks, while in code-improvement tasks it may inspect the full visible test suite.

\subsubsection{Executor}

\textbf{Purpose.} The Executor implements the candidate modification proposed for the current round.

\textbf{Inputs.} Investigation artifact, accepted system state, task specification, and the set of files, parameters, or components that are admissible for modification.

\textbf{Outputs.} A concrete candidate system state for the current round, together with the modified artifacts needed for evaluation.

\textbf{Constraints.} The Executor is limited to the admissible action space defined by the task. Depending on the skill, this may include prompt files, hyperparameter configurations, rules, or source code. The Executor does not perform final review and may not modify protected evaluation data or forbidden components such as fixed model architectures.

\subsubsection{Reviewer}

\textbf{Purpose.} The Reviewer evaluates the candidate system state and determines whether it should be accepted.

\textbf{Inputs.} Candidate system state, current accepted state, evaluation interface, primary metric, and task-specific decision criteria.

\textbf{Outputs.} Evaluation results, round-to-round deltas, and a structured verdict such as accept, reject, retry, or terminate.

\textbf{Constraints.} The Reviewer is separated from both hypothesis generation and implementation. It does not propose new changes or modify the candidate directly. Depending on the task, the Reviewer may be implemented as a metric-driven evaluator or as an LLM-based critic, but in both cases it must apply the predefined evaluation criteria rather than ad hoc preferences.

\subsection{Example Output Directory Structure}
\label{app:output-structure}

Figure~\ref{fig:output-structure} shows an example directory layout for an EPOCH task and its run-level artifacts. The structure reflects the protocol's emphasis on baseline establishment, per-round tracking, and final run summarization. Concrete tasks may add task-specific files, but the same high-level organization is preserved across protocol instantiations.

\begin{figure}[t]
\centering
\begin{minipage}{0.9\linewidth}
\begin{Verbatim}[frame=single,fontsize=\small]
projects/
├── <slug>_run.yaml          # EPOCH task specification
└── <slug>/                  # task directory
    ├── evaluate.py          # evaluation entry point
    ├── src/                 # prompts, code, or model config
    ├── tests/               # code-improvement test suite
    ├── rules/               # rule-based artifacts
    └── <run_id>/            # run-level artifacts
        ├── baseline_metrics.json
        ├── investigation_report_round_N.md
        ├── proposed_metrics_round_N.json
        ├── delta_round_N.json
        └── run_summary.md
\end{Verbatim}
\end{minipage}
\caption{Example EPOCH task and run artifact directory structure.}
\label{fig:output-structure}
\end{figure}

\subsection{Generic EPOCH Task Specification}
\label{app:generic-config}

Figure~\ref{fig:epoch-template} shows a compact generic task specification for EPOCH. The specification defines the optimization goal, protocol budget, enabled phases, active roles, evaluation interface, and tracking backend. Concrete task instantiations specialize this template by selecting a task type and filling in task-specific fields such as data access constraints, model configuration, and admissible modification space.

\begin{figure}[t]
\centering
\begin{minipage}{0.95\linewidth}
\begin{Verbatim}[frame=single,fontsize=\small]
project:
  name: "<project name>"
  slug: "<project_slug>"

run:
  goal: "<optimization goal>"
  task_type: "<prompt_tune|finetune|rule_based|code_improvement|custom>"
  max_rounds: <int>
  max_retries_per_round: <int>

phases:
  baseline_construction: <true|false>
  multi_round_optimization: <true|false>

roles:
  seed_planner: <true|false>
  baseline_executor: <true|false>
  orchestrator: true
  investigator: <true|false>
  executor: <true|false>
  reviewer:
    enabled: <true|false>
    mode: "<metric_driven |llm_critic>"

data:
  source: "<dataset|local path|api>"
  train_split: "<train split|null>"
  eval_split: "<eval split|null>"

model:
  type: "<llm|ml_model|rule_system|code>"
  name: "<model name|null>"

investigation:
  samples: <int>
  access_scope: "<train_only|eval_only|full_visible_tests|custom>"

evaluation:
  primary_metric: "<metric>"
  min_delta: <float>
  deterministic: <true|false>
  train_cmd: "<command|null>"
  eval_cmd: "<command>"
  acceptance_rule: "<text description>"

tracking:
  backend: "<local_logs|structured_files|github_prs|custom>"

git:
  enabled: <true|false>
  target_branch: "<branch|null>"
\end{Verbatim}
\end{minipage}
\caption{Compact generic EPOCH task specification.}
\label{fig:epoch-template}
\end{figure}

Not all fields are required for every task. For example, code-improvement tasks may use visible tests instead of train/eval splits, while prompt-tuning tasks may require additional LLM configuration and leakage constraints. The protocol remains fixed at the orchestration level, while task-specific behavior is specialized through this configuration interface.

\end{document}